\newif\ifsubmission
\newif\ifappendix

\submissionfalse

\appendixtrue

\documentclass[letterpaper]{article} 
\ifsubmission
\usepackage[submission]{aaai2026}
\else
\usepackage{aaai2026} 
\fi

\usepackage{times} 
\usepackage{helvet} 
\usepackage{courier} 
\usepackage[hyphens]{url} 
\usepackage{graphicx} 
\usepackage{natbib} 
\usepackage{caption} 
\usepackage{algorithm}
\usepackage{algpseudocode}

\usepackage{booktabs}
\usepackage{multirow}
\usepackage{comment}

\usepackage{amsmath}
\usepackage{amssymb}
\usepackage{bm}
\usepackage{cancel}
\usepackage{mathtools}

\usepackage{newfloat}
\usepackage{listings}
\DeclareCaptionStyle{ruled}{labelfont=normalfont,labelsep=colon,strut=off} 
 \floatstyle{ruled} \newfloat{listing}{tb}{lst}{}
\floatname{listing}{Listing}
\usepackage{color} 

\newcommand{\evals}{\textsc{Evaluation}}
\newcommand{\SO}{\textsc{Seq. Opt.}}
\newcommand{\SOfull}{Sequential Optimization}
\newcommand{\CBE}{\textsc{C-BE}} 
\newcommand{\DBE}{\textsc{D-BE}} 
\newcommand{\CBEfull}{Coupled Updates with Batched Evaluation}
\newcommand{\DBEfull}{Decoupled Updates with Batched Evaluation}

\newcommand{\argmax}{\mathop{\mathrm{argmax}}}
\newcommand{\xvec}{\mathbf{x}}
\newcommand{\Xvec}{\mathbf{X}}
\newcommand{\asum}{\alpha_{\mathrm{sum}}}

\definecolor{cbe}{HTML}{1F77B4}  
\definecolor{dbe}{HTML}{D62728}  
\newcommand{\hlcbe}[1]{\begingroup\setlength{\fboxsep}{1pt}\colorbox{cbe!12}{#1}\endgroup}
\newcommand{\hldbe}[1]{\begingroup\setlength{\fboxsep}{1pt}\colorbox{dbe!12}{#1}\endgroup}
\newcommand{\cbe}[1]{\textcolor{cbe}{#1}}
\newcommand{\dbe}[1]{\textcolor{dbe}{#1}}

\algtext*{EndFor}
\algtext*{EndWhile}
\algtext*{EndIf}
\algtext*{EndProcedure}
\algtext*{EndFunction}
\algnewcommand{\LineComment}[1]{\State \(\triangleright\) #1}

\title{Batch Acquisition Function Evaluations and Decouple Optimizer Updates for Faster Bayesian Optimization}

\author{
    Kaichi Irie \textsuperscript{\rm 1,2,3}, 
    Shuhei Watanabe \textsuperscript{\rm 2,3},
    Masaki Onishi \textsuperscript{\rm 3}
}
\affiliations{ 
    \textsuperscript{\rm 1}Graduate School of Informatics, Kyoto University\\
    \textsuperscript{\rm 2}Preferred Networks Inc.\\ \textsuperscript{\rm 3}Advanced Industrial Science and Technology (AIST)\\
    irie.kaichi.24x@st.kyoto-u.ac.jp
    }

\begin{document}
    \maketitle
    \begin{abstract}
Bayesian optimization (BO) efficiently finds high-performing parameters by maximizing an acquisition function, which models the promise of parameters.
A major computational bottleneck arises in acquisition function optimization, where multi-start optimization (MSO) with quasi-Newton (QN) methods is required due to the non-convexity of the acquisition function.
BoTorch, a widely used BO library, currently optimizes the summed acquisition function over multiple points, leading to the speedup of MSO owing to PyTorch batching.
Nevertheless, this paper empirically demonstrates the suboptimality of this approach in terms of off-diagonal approximation errors in the inverse Hessian of a QN method, slowing down its convergence.
To address this problem, we propose to decouple QN updates using a coroutine while batching the acquisition function calls.
Our approach not only yields the theoretically identical convergence to the sequential MSO but also drastically reduces the wall-clock time compared to the previous approaches.
Our approach is available in \texttt{GPSampler} in Optuna, effectively reducing its computational overhead.
    \end{abstract}

    \ifsubmission
    \begin{links}
        \link{Code}{https://anonymous.4open.science/r/aaai2026-anon-exp-4795} 
    \end{links}
    \else
    \begin{links}
        \link{Code}{https://github.com/Kaichi-Irie/faster-batched-acqf-opt-experiments/tree/submission}
    \end{links}
    \fi

    \section{Introduction}
\label{sec:intro}

Bayesian optimization (BO) is a prevalent method to reduce trial-and-error iterations in many applications with an expensive objective such as materials discovery~\cite{xue2016accelerated,li2017rapid,vahid2018new}, hyperparameter optimization (HPO) of machine learning models~\cite{loshchilov2016cma, chen2018bayesian, feurer2019automated}, and drug discovery~\cite{schneider2020rethinking}.
Since the acquisition function optimization plays a central role in an efficient search, multi-start optimization (MSO) by a quasi-Newton (QN) method, which is essential to enhance the solution quality, is employed by BoTorch~\cite{balandat2020botorch} and Optuna~\cite{akiba2019optuna}.

Although MSO is widely regarded as the gold standard, its overhead can be non-negligible when each optimization is run sequentially (denoted \SO), as it multiplies acquisition-function evaluations that each entail an expensive Gaussian-process regressor call.
BoTorch optimizes the sum of acquisition-function values across multiple restarts~\footnote{See ``Multiple Random Restarts'' in the BoTorch documentation. URL: \url{https://botorch.org/docs/v0.14.0/optimization/#multiple-random-restarts}.} to mitigate this cost.
This formulation enables PyTorch to batch acquisition function evaluations, reducing overall overhead while preserving the theoretically exact per-point gradients.


Despite these advantages, this formulation substantially slows down the convergence of QN methods based on our experiments.
We empirically show the slowdown is explained by approximation errors in cross derivatives or off-diagonal blocks, dubbed \emph{off-diagonal artifacts}, between dimensions of independent points.
Off-diagonal artifacts distort search directions, leading to slower convergence.

As a remedy, we propose \textbf{D}ecoupling QN updates per restart while \textbf{B}atching acquisition function \textbf{E}valuations (\DBE).
By contrast, we refer to the BoTorch practice, which \textbf{C}ouples QN updates while using \textbf{B}atched \textbf{E}valuations, as \CBE.
Note that the BoTorch v0.15.0 changed from \CBE\ to \DBE\ (, although it does not use a coroutine) independently of this work.
To our knowledge, there is no practical block-structure-aware, bound-constrained, which is ubiquitous in BO, QN algorithm that removes these off-diagonal artifacts.
\DBE\ sidesteps this difficulty by keeping per-point states independently while retaining batched evaluations, by combining a coroutine and batching---no new solver is required.
In experiments, \DBE\ eliminates off-diagonal artifacts while leveraging hardware throughput and preserving solution quality, yielding up to $1.5\times$ and $1.1\times$ wall-clock speedups over \SO\ and \CBE, respectively.


In summary, our contributions are as follows:
\begin{enumerate}
    \item Present a pitfall of \CBE\ that slows the convergence, focusing on off-diagonal artifacts, 
    \item Propose \DBE\ achieved by a novel combination of a coroutine and batching, which avoids the pitfall without degrading solution quality, and
    \item Demonstrate that \DBE\ is faster than \SO\ and \CBE\ in experiments.
\end{enumerate}
Notice that our approach has been successfully merged into \texttt{GPSampler} in Optuna, significantly accelerating its runtime.


    \begin{figure*}[t]
  \centering
  \includegraphics[width=0.8\textwidth]{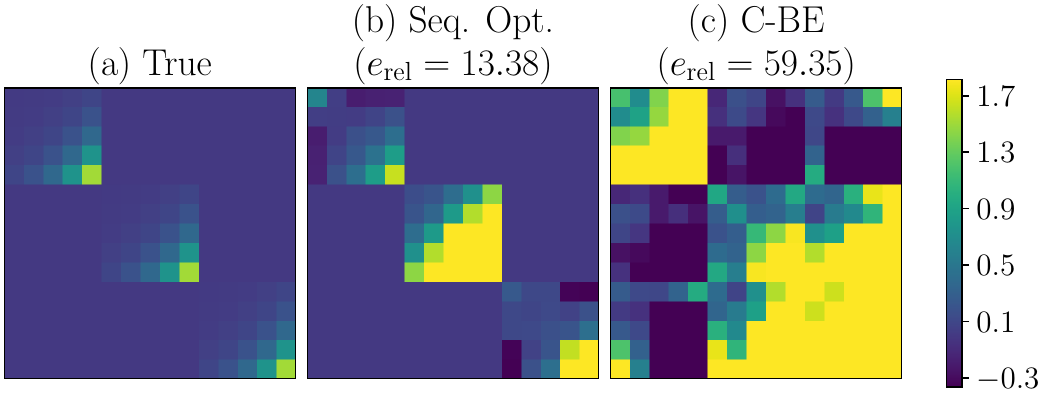}
  \caption{
  Contour maps of the inverse Hessian (\textbf{Left}) and the inverse Hessians approximated by L-BFGS-B with \SO\ (\textbf{Center}) and \CBE\ (\textbf{Right}) on the Rosenbrock function, evaluated near the constrained minimizer ($B = 3, D = 5, \xvec \in [0, 3]^D$).
  Each figure has $15 \times 15$ tiles, and the $(i, j)$-th tile corresponds to the colormap for the $(i,j)$-th element of the (approximated) inverse Hessian.
  Blue and yellow represent lower and higher values, respectively.
  Each subtitle reports $e_{\mathrm{rel}}(H)=\|H-H_{\mathrm{true}}\|_{F}/\|H_{\mathrm{true}}\|_{F}$.
  \textbf{Left}: The true inverse Hessian exhibits zero at off-diagonal blocks.
  \textbf{Center}: The approximated inverse Hessian by \SO.
  Off-diagonal blocks show zero everywhere.
  \textbf{Right}: The approximated inverse Hessian by \CBE.
  Off-diagonal blocks are dense because \CBE\ does not allow QN methods to be aware of the zero off-diagonal block nature.
  }
  \label{fig:hessian_mismatch}
\end{figure*}

\section{Background}
\label{sec:background}


\subsection{Bayesian Optimization}
Given a function $f(\xvec): \mathbb{R}^D \rightarrow \mathbb{R}$ to be maximized, Bayesian optimization iteratively suggests a parameter vector $\xvec \in \mathbb{R}^D$ by maximizing an acquisition function $\alpha: \mathbb{R}^D \rightarrow \mathbb{R}$.
When the acquisition function is defined on a continuous space, QN methods are widely used for the optimization owing to their fast convergence.
Even if the analytical expression of gradients $\nabla \alpha(\xvec)$ is cumbersome to derive, PyTorch enables us to compute the gradients numerically using automatic differentiation (AD)~\cite{ament2023unexpected,daulton2020differentiable}.
Although QN methods achieve fast convergence, the non-convexity of acquisition functions necessitates MSO.
Namely, the acquisition function is optimized $B$ times independently to enhance the solution quality.
The goal of this paper is to propose a method to speed up the $B$ times of independent optimizations without degrading the solution quality. 



\subsection{Quasi-Newton (QN) Methods}

QN methods, such as BFGS, L-BFGS, and L-BFGS-B, optimize a function using first-order gradients together with an approximation to the inverse Hessian, i.e., the matrix of second derivatives~\cite{nocedal1980updating, liu1989limited, byrd1994representations, byrd1995limited}.
In particular, L-BFGS-B is widely used for optimizing acquisition functions because of its fast convergence and ability to handle box constraints.


\subsection{Multi-Start Optimization: Sequential vs. Batching}

Two schemes have been considered for MSO so far, (1) \SOfull\ (\SO), which performs an independent optimization from each starting point sequentially (see Algorithm 2 in Appendix A for more details), and (2) \CBEfull\ (\CBE).
More concretely, \CBE\ optimizes the summation of acquisition function values from independent restarts.
Namely, \CBE\ optimizes the following function with QN methods:
\begin{equation}
  \asum(\Xvec) \coloneqq \sum_{b=1}^{B}\alpha\big(\xvec^{(b)}\big),
  \label{eq:coupled-objective}
\end{equation}
where $\xvec^{(b)} \in \mathbb{R}^D$ is a point in the $b$-th optimization and $\Xvec =\{\xvec^{(b)}\}_{b=1}^B \in\mathbb{R}^{B\times D}$ is a matrix with each point from the $b$-th optimization at the $b$-th column.
Since $\asum$ is additively separable, the gradient of $\asum$ with respect to $\xvec^{(b)}$ matches that of $\alpha(x^{(b)})$, enabling a batched evaluation to get the acquisition function values and their gradients at multiple points simultaneously.
In a nutshell, \CBE\ optimizes a $BD$-dimensional (additively separable) function instead of solving a $D$-dimensional function $B$ times independently.
BoTorch uses this technique to accelerate MSO.

    \section{Pitfall of \CBE: Off-Diagonal Artifacts}
\label{sec:cbe}

Although \CBE\ has been used to accelerate Bayesian optimization, it can in fact slow down each BO iteration or degrade the solution quality when it is used with QN methods.

When using first-order (gradient-based) methods, each block of $\nabla_{\Xvec}\asum(\Xvec)$ equals the per-restart gradient $\nabla\alpha(\xvec^{(b)})$, making the convergence of \CBE\ comparable to that of \SO\
By contrast, QN methods approximate the inverse Hessian, and the accuracy of this approximation strongly affects the convergence rate.
Importantly, the Hessian (and its inverse) of the summed acquisition over multiple restarts has zero cross-partial derivatives between variables associated with different restarts.
More precisely, the true Hessian of $\asum$ is block-diagonal:
\begin{equation}
  H \coloneqq \nabla_{\Xvec}^{2}\asum(\Xvec) = \begin{bmatrix}
    \nabla^{2}\alpha(\xvec^{(1)}) & & \mathbf{0} \\
    & \ddots & \\
    \mathbf{0} & & \nabla^{2}\alpha(\xvec^{(B)})
  \end{bmatrix}.
\end{equation}

However, structure-oblivious QN updates in the $BD$-dimensional space typically maintain a dense inverse-Hessian approximation, injecting non-zero values into off-diagonal blocks and distorting search directions.
We refer to this phenomenon as \emph{off-diagonal artifacts}.
Figure~\ref{fig:hessian_mismatch} visualizes the inverse Hessian and its L-BFGS-B approximation (memory size $m = 10$) under \SO\ and \CBE.
The figure intuitively shows \CBE\ induces off-diagonal artifacts.

\begin{figure}[t]
  \centering
  \includegraphics[width=0.95\columnwidth]{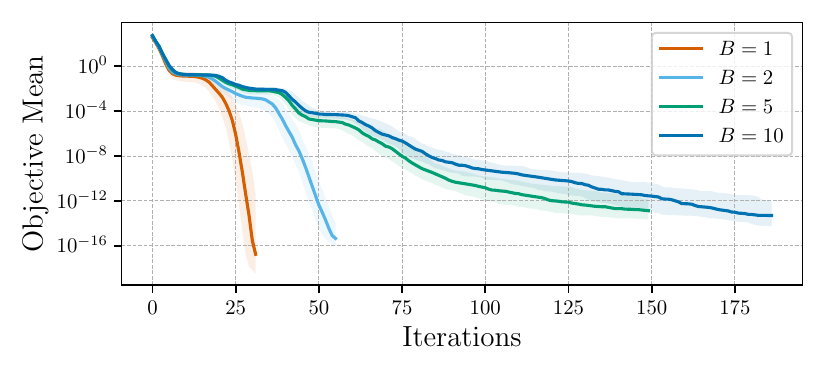}
  \caption{
  Convergence speed of \CBE\ when using L-BFGS-B with the memory size $m=10$ on the Rosenbrock function ($D=5, \xvec \in [0,3]^D$).
  The figure shows the objective mean over $B$ restarts at each iteration.
  Each optimization is repeated $1000 / B$ times.
  Each solid line and weak-color band represents the median and the $\pm$ IQR of the objective mean over $1000 / B$ runs, respectively.
  \SO\ corresponds to $B=1$.
  As the number of restarts $B$ increases, the convergence of \CBE\ requires substantially more iterations.
  }
  \label{fig:convergence_rosenbrock}
\end{figure}

The next question is: ``How much do these off-diagonal artifacts affect the convergence speed?''.
Figure~\ref{fig:convergence_rosenbrock} answers this question using the Rosenbrock function.
Looking at the medians, the optimization reaches an objective value of $10^{-12}$ in $\sim 30$ iterations for \SO\ ($B=1$), whereas \CBE\ with $B=2$ and $B \in \{5,10\}$ requires $\sim 50$ and more than $120$ iterations, respectively.
Overall, MSO from $10$ initial points would require evaluations at about $300~(= 30 \times 10)$ points for $B = 1$ and more than $1200 = (120 \times 10)$ for $B = 10$, concluding that a na\"ive \CBE\ weakens the speedup effect.
This is also the case for BFGS as shown in Figures 3--5 of Appendix B.
It is worth noting that there is no principled way for L-BFGS-B to handle this phenomenon, although BFGS and L-BFGS have been extended to do so by \citeauthor{griewank1982partitioned} and \citeauthor{bigeon2023framework}, respectively.
Namely, no work has addressed this issue for the acquisition function optimization, where each parameter has its bounds.
The next section introduces a simple yet novel remedy that enables block-diagonal-structure-aware speedups without requiring new QN solvers.

    \begin{algorithm}[t]
  \caption{MSO: \CBE\ vs.\ \DBE\ }
  \label{alg:cbe-vs-dbe}
  \begin{algorithmic}[1]
    \Statex{\textbf{Inputs}: $\alpha$ (Acquisition function), $B$ (number of restarts), $\Xvec_0=\{\xvec^{(b)}_0\}_{b=1}^{B}$ (initial points)}
    \Statex{\textbf{Output}: $\xvec^\star
\in \argmax_{\xvec} \alpha(\xvec)$}
    \State{$\Xvec\leftarrow \Xvec_0$}
    \State{\hlcbe{\cbe{[C-BE]}}~\hlcbe{Initialize a QN optimizer $O_\text{CBE}$ on $\Xvec\in\mathbb{R}^{B\times D}$}}
    \State{\hldbe{\dbe{[D-BE]}}~\hldbe{Initialize independent QN optimizers $O_1,\dots,O_B$}}
    \While{not converged}
      \LineComment{Batched Evaluation}
      \State{$\mathbf{a}\leftarrow \{\alpha(\xvec^{(b)})\}_{b=1}^B$, $\mathbf{g}\leftarrow \{\nabla\alpha(\xvec^{(b)})\}_{b=1}^B$} 
      \LineComment{QN Updates}
      \State{\hlcbe{\cbe{[\CBE]}}}
      \State{\hlcbe{$\asum \leftarrow \sum_{b=1}^{B} \alpha^{(b)} \coloneqq \sum_{b=1}^B \alpha(\xvec^{(b)})$}} 
      \State{\hlcbe{$\Xvec\leftarrow O_{\mathrm{CBE}}(\Xvec, \asum, \mathbf{g})$}}
      \State{\hldbe{\dbe{[\DBE]}}}
      \State{\hldbe{\textbf{for} $b=1, \dots, B$ \textbf{do}}}
      \State{\hldbe{\quad $\xvec^{(b)}\leftarrow O_b(\xvec^{(b)}, \alpha^{(b)}, \mathbf{g}^{(b)})$}}
    \EndWhile
    \Return{$\xvec^\star = \argmax_{\xvec \in \{\xvec^{(b)}\}_{b=1}^{B}} \alpha(\xvec)$}
  \end{algorithmic}
\end{algorithm}

\section{Decoupling Updates Converge Faster without Difficult Mathematics}
\label{sec:dbe}


\subsection{Why Is \DBE\ Better Than \CBE?}

Algorithm~\ref{alg:cbe-vs-dbe} highlights the differences between \DBE\ (our proposal) and \CBE.
\DBE\ is characterized by independent initialization and per-restart updates.
Concretely, after each batched evaluation, \DBE\ performs $B$ independent L-BFGS-B updates---one per restart---using only its own history, thereby avoiding the off-diagonal artifacts present in \CBE.

Batching benefits not only acquisition-function evaluations but also the L-BFGS-B updates; however, the gain for the latter is limited.
We now explain why.
For example, a single evaluation of the expected improvement using a Gaussian process regressor with $n$ training data points costs $\mathcal{O}(n^2+nD)$, so evaluating $B$ points costs $\mathcal{O}(B(n^2+nD))$.
By contrast, the L-BFGS-B update with the memory size $m$ for $B$ independent optimizations costs $\mathcal{O}(BmD)$.
In regimes where $n \gg m$ (typically after tens to hundreds of trials, with $m\in[5,20]$), it is cost-effective to batch only evaluations while keeping each optimizer independent.
As a result, \DBE\ avoids the off-diagonal artifacts induced by \CBE.

Most importantly, each restart in \DBE\ theoretically reproduces the per-restart update trajectory as in \SO\ under the identical initialization and termination policies.
Notice, however, that AD yields slightly different gradients between a batched evaluation and a single evaluation due to modulo floating-point nondeterminism, so the trajectory may not be identical.
The general form of \DBE\ maintains a set of ongoing optimization indices $\mathcal{A}\subseteq\{1,\dots,B\}$, and prunes the converged optimizations.
This mechanism allows \DBE\ to progressively shrink the batch size, reducing additional computational overhead.
Note that Algorithm~\ref{alg:cbe-vs-dbe} does not detail the point-wise termination owing to brevity.
In contrast, \CBE\ employs a QN optimizer with a shared QN state over all the restarts.
Since detaching the converged optimizations is not feasible in \CBE, the overall runtime unnecessarily inflates.


\subsection{Decouple L-BFGS-B Updates by Coroutine}

We have discussed the benefit of using \DBE\ instead of \CBE.
However, decoupling the L-BFGS-B updates often requires modifying the L-BFGS-B implementation directly.
For example, SciPy provides L-BFGS-B while per-QN-iteration hooks are not exposed, making batched evaluations for independent optimizations non-trivial.
Even in this scenario, we do not have to implement our own QN methods if we use a coroutine library.
A coroutine is a special type of function that can pause its execution and be resumed later from the same point, allowing for cooperative multitasking and simplifying asynchronous programming.
\DBE\ can be achieved by using a coroutine, which is composed of (1) one batch evaluator, and (2) $B$ workers for each independent optimization.
In the coroutine, the evaluator performs a batched evaluation, it dispatches $(\alpha^{(b)},g^{(b)})$ to each worker, and each worker updates its QN state.
Repeating this coroutine enables \DBE\ without any new solvers.

    \section{Benchmarking Experiments}
\label{sec:experiments}

We finally conduct the speedup effect and the solution quality check of \DBE\ against \SO\ and \CBE\ through a BO application.
The Rastrigin function from COCO (BBOB) suite~\cite{hansen2009real, hansen2021coco} available in OptunaHub~\cite{ozaki2025optunahub} is taken as an objective function for BO.
The benchmarking results on other functions are available in Table 2 of Appendix.
Note that the optimization is performed in the minimization direction in this section.
Each BO iteration first fits a Gaussian process regressor with a Mat\'{e}rn-$\nu{=}5/2$ kernel on a set of observed pairs of a parameter vector and the corresponding objective value.
Then, the next parameter vector is determined based on log expected improvement (LogEI).
LogEI is optimized by MSO using L-BFGS-B in each BO iteration.



\begin{table}[t]
  \centering
  \small
  \begin{tabular}{lcccc}
    \toprule \textbf{$D$}        & \textbf{Method} & \textbf{Best Value} $\downarrow$ & \textbf{Runtime (s)} $\downarrow$ & \textbf{Iters.} $\downarrow$ \\
    \midrule \multirow{3}{*}{5}  & \SO             & $10.94$                          & $92.7$                            & $\mathbf{11.0}$              \\
                                 & \CBE            & $11.28$                          & $62.0$                            & $35.0$                       \\
                                 & \DBE            & $\mathbf{10.85}$                 & $\mathbf{59.2}$                   & $\mathbf{11.0}$              \\
    \midrule \multirow{3}{*}{10} & \SO             & $28.24$                          & $96.5$                            & $\mathbf{14.8}$              \\
                                 & \CBE            & $24.71$                          & $73.7$                            & $54.2$                       \\
                                 & \DBE            & $\mathbf{23.89}$                 & $\mathbf{67.3}$                   & $15.0$                       \\
    \midrule \multirow{3}{*}{20} & \SO             & $59.08$                          & $129.4$                           & $21.5$                       \\
                                 & \CBE            & $57.04$                          & $95.4$                            & $68.8$                       \\
                                 & \DBE            & $\mathbf{55.81}$                 & $\mathbf{86.4}$                   & $\mathbf{20.0}$              \\
    \midrule \multirow{3}{*}{40} & \SO             & $88.31$                          & $169.2$                           & $26.5$                       \\
                                 & \CBE            & $100.36$                         & $120.6$                           & $87.0$                       \\
                                 & \DBE            & $\mathbf{76.48}$                 & $\mathbf{108.2}$                  & $\mathbf{25.8}$              \\
    \midrule
  \end{tabular}
  \caption{
    The benchmarking results of BO with $300$ trials using L-BFGS-B with the memory size of $m=10$ and $B=10$ restarts on the $D$-dimensional Rastrigin function ($D=5,10,20,40$).
    The termination criterion of L-BFGS-B is $200$ iterations or $\|\nabla \alpha(\mathbf{x})\|_{\infty}\le 10^{-2}$.
    Each column shows the median over $20$ independent runs, each with a different random seed.
    \textbf{Best Value}: the minimum objective value among the 300 trials minus the best objective value over all runs.
    \textbf{Runtime (s)}: the total wall-clock time for BO.
    \textbf{Iters.}: the median of the L-BFGS-B iteration count over $300$ Trials $\times 10$ restarts. Lower is better.
  }
  \label{tab:bo_benchmark_rastrigin}
\end{table}

Table~\ref{tab:bo_benchmark_rastrigin} shows the benchmarking results.
Based on the results, all methods achieved comparable (median) final objective values under the shared iteration cap.
As can be seen in the Iters. column, the L-BFGS-B iteration count drastically increased in \CBE.
For example, \CBE\ required substantially more L-BFGS-B iterations than \SO\ as $D$ grew, e.g., $68.8/21.5\!\approx\!3.2$ for $D=20$ and $87.0/26.5\!\approx\!3.3$ for $D{=}40$.
This observation is consistent with the discussion regarding the off-diagonal artifacts.
These extra iterations can offset batching gains, making \CBE's wall-clock time comparable to---and on some objectives worse than---\SO\
By contrast, \DBE\ matches the iteration counts of the \SO\ baseline while exploiting batched gradient evaluations, having reduced the runtime of the acquisition function optimization ($\sim\!1.5\times$ faster than \SO).
The similar pattern holds on \textit{Sphere}, \textit{Attractive Sector} and \textit{Step Ellipsoidal} from the COCO (BBOB) suite~\cite{hansen2009real, hansen2021coco};
See Appendix~\ref{app:detailed-results} (Table~\ref{tab:bo_benchmark_all}) for more details.
Note that while \CBE\ occasionally outperformed \SO\ for $D\in\{10, 20\}$, we attribute this to the difference in the convergence behavior rather than a systematic benefit of coupling.
More specifically, the acquisition function optimization for \CBE\ is mostly terminated by the gradient tolerance but not the iteration count, meaning that the optimization results for \CBE\ are not premature.
This leads to very different optimization results between \CBE\ and \SO, making it possible for \CBE\ to sometimes outperform \SO

    \section{Conclusion}
\label{sec:conclusion}

This paper discussed the speedup effect of BO in light of the acquisition function optimization.
Although the optimization of the summed acquisition function at multiple points has been known to be a fast implementation for MSO, we empirically showed that either the convergence speed or the solution quality is degraded owing to off-diagonal artifacts.
In principle, \CBE\ degrades the solution quality when the iteration count is capped, or it takes significantly longer when we wait until the solution quality is satisfied.
Mathematically speaking, a variant of L-BFGS-B can optimize efficiently as is, but such an extension is challenging.
To this end, we proposed an explicit decoupling approach to circumvent the off-diagonal artifacts.
In general, such decoupling requires L-BFGS-B to take a specific design especially when the BO implementation relies on PyTorch owing to its incompatibility with multi-processing.
Meanwhile, our proposition only requires a coroutine to realize the decoupling.
The experiments demonstrated that our approach accelerated BO while keeping the original performance.




    \newpage
    \bibliography{batched_eval}

    \ifappendix
    \clearpage
    \appendix
    \suppressfloats[t]   
    \section*{Appendix}
    \section{\SOfull\ (\SO)}
\label{app:so} 
\begin{algorithm}[t]\small
  \caption{MSO: \SO}
  \label{alg:so-appendix}
  \begin{algorithmic}[1]
    \Statex{\textbf{Inputs}: $\alpha$ (Acquisition function), $B$ (number of restarts), $\xvec_0=\{\xvec^{(b)}_0\}_{b=1}^{B}$ (initial points)}
    \Statex{\textbf{Output}: $\xvec^\star
\in \argmax_{\xvec} \alpha(\xvec)$}
    \State{$\xvec\leftarrow \xvec_0$}
    \For{$b=1,\dots,B$}
      \State{Initialize QN optimizer $O_b$; set $\mathbf{x}^{(b)} \leftarrow \mathbf{x}^{(b)}_{0}$}
      \While{restart $b$ not converged}
        \LineComment{Evaluation (no batching across restarts)}
        \State{$a^{(b)} \leftarrow 
        \alpha(\mathbf{x}^{(b)})$;\quad $\mathbf{g}^{(b)} \leftarrow \nabla \alpha(\mathbf{x}^{(b)})$} 
        \LineComment{QN Updates}
        \State{$\mathbf{x}^{(b)} \leftarrow O_b\big(\mathbf{x}^{(b)},\, a^{(b)},\, \mathbf{g}^{(b)}\big)$}
      \EndWhile
    \EndFor
    \Return{$\xvec^\star = \argmax_{\{\xvec^{(b)}\}_{b=1}^{B}} \alpha(\xvec)$}
  \end{algorithmic}
\end{algorithm}

Algorithm~\ref{alg:so-appendix} presents the \SO\ algorithm, which serves as a baseline.
Each restart maintains its own optimizer state, and no batching is performed across restarts.
As a result, no cross-restart information is shared, and the per-restart curvature is preserved by construction.

    \section{Off-Diagonal Artifact Effect on BFGS}
\label{app:bfgs}

\begin{figure}[t]
  \centering
  \includegraphics[width=0.95\columnwidth]{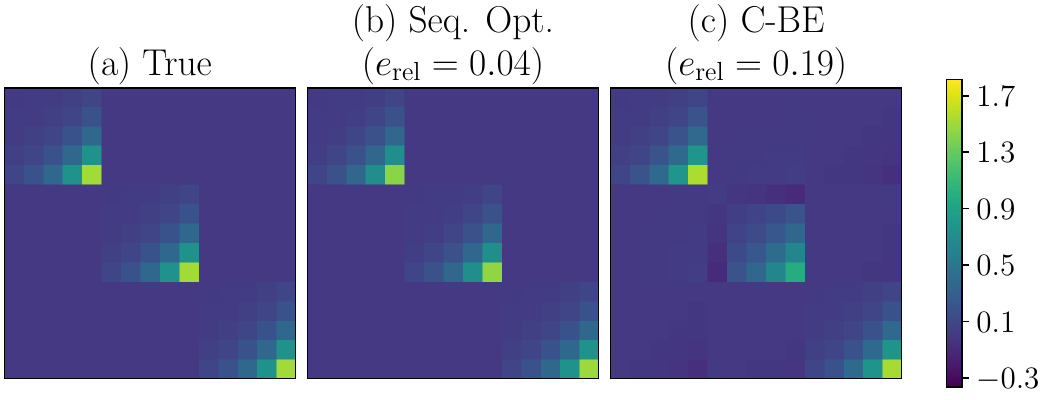}
  \vspace{-2mm}
  \caption{
    Contour maps of the inverse Hessian (\textbf{Left}) and its approximations by \SO\ (\textbf{Center}) and \CBE\ (\textbf{Right}).
    This figure follows the same setup as Figure~\ref{fig:hessian_mismatch} except that BFGS, i.e., the memory is not limited, is used instead of L-BFGS-B.
  }
  \label{fig:hessian_mismatch_bfgs_b3}
\end{figure}

\begin{figure}[t]
  \centering
  \includegraphics[width=0.95\columnwidth]{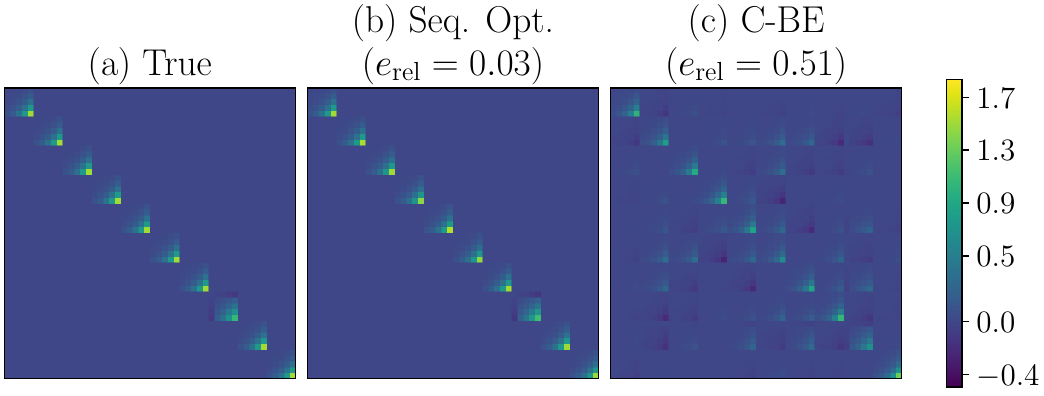}
  \vspace{-2mm}
  \caption{
    Contour maps of the inverse Hessian (\textbf{Left}) and its approximations by \SO\ (\textbf{Center}) and \CBE\ (\textbf{Right}).
    This figure follows the same setup as Figure~\ref{fig:hessian_mismatch} except that BFGS, i.e., the memory is not limited, is used instead of L-BFGS-B, and $B=10$ is used instead of $B=3$.
    Off-diagonal artifacts are more prominent for a large $B$.
  }
  \label{fig:hessian_mismatch_bfgs_b10}
\end{figure}

\begin{figure}[t]
  \centering
  \includegraphics[width=0.95\columnwidth]{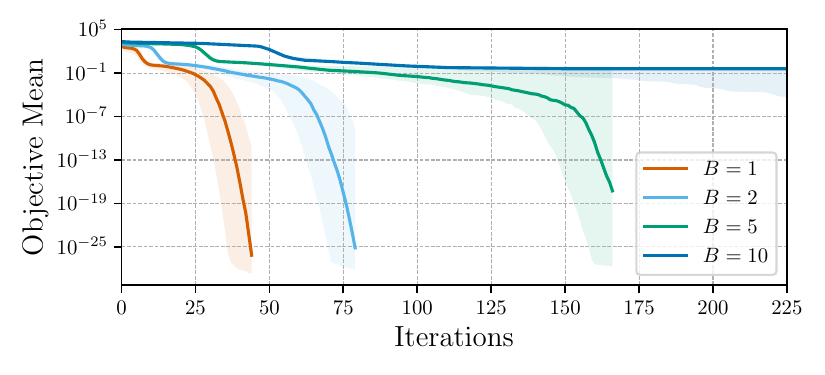}
  \vspace{-2mm}
  \caption{
    Convergence speed of \CBE\ when using BFGS.
    This figure follows the same setup as Figure~\ref{fig:convergence_rosenbrock} except that BFGS is used instead of L-BFGS-B.
    As the number of restarts $B$ increases, the convergence of \CBE\ requires substantially more iterations.
  }
  \label{fig:convergence_rosenbrock_bfgs_d5}
\end{figure}

This section verifies that the observed degradation under \CBE\ does not stem from limiting the memory of BFGS.
The setup in this section follows Section~\ref{sec:cbe}.
Figures~\ref{fig:hessian_mismatch_bfgs_b3} and \ref{fig:hessian_mismatch_bfgs_b10} show that \CBE\ injects off-diagonal mass into the inverse Hessian approximation, which should be zero, while \SO\ preserves the block-diagonal structure.
Consistently, iteration counts increase markedly under \CBE\ relative to \SO\ as seen in Figure~\ref{fig:convergence_rosenbrock_bfgs_d5}.
These observations confirm that the off-diagonal artifacts are caused by coupling QN updates for a separable problem rather than by limiting the memory.

    \section{Full Benchmarking Results}
\label{app:detailed-results}

\begin{table*}[t]
  \centering
  \small
  \setlength{\tabcolsep}{1mm}

  \begin{tabular}{lcccc}
    \toprule \textbf{Objective}        & \textbf{Method} & \textbf{Best Value} $\downarrow$ & \textbf{Runtime (s)} $\downarrow$ & \textbf{Iters.} $\downarrow$ \\
    \midrule \multirow{3}{*}{5$D$ Sphere}     & \SO             & $4.026\times 10^{-5}$            & $196.9$                           & $\mathbf{7.0}$               \\
                                             & \CBE            & $\mathbf{1.068\times 10^{-5}}$   & $\mathbf{121.8}$                  & $7.5$                        \\
                                             & \DBE            & $5.167\times 10^{-5}$            & $128.2$                           & $\mathbf{7.0}$               \\
    \midrule \multirow{3}{*}{10$D$ Sphere}    & \SO             & $9.632\times 10^{-4}$            & $213.6$                           & $13.0$                       \\
                                             & \CBE            & $\mathbf{1.846\times 10^{-4}}$   & $146.8$                           & $26.0$                       \\
                                             & \DBE            & $4.973\times 10^{-4}$            & $\mathbf{141.6}$                  & $\mathbf{12.8}$              \\
    \midrule \multirow{3}{*}{20$D$ Sphere}    & \SO             & $8.471\times 10^{-3}$            & $225.4$                           & $15.5$                       \\
                                             & \CBE            & $9.820\times 10^{-3}$            & $173.1$                           & $48.8$                       \\
                                             & \DBE            & $\mathbf{7.246\times 10^{-3}}$   & $\mathbf{147.4}$                  & $\mathbf{15.0}$              \\
    \midrule \multirow{3}{*}{40$D$ Sphere}    & \SO             & $0.04$                           & $291.4$                           & $\mathbf{21.2}$              \\
                                             & \CBE            & $0.03$                           & $220.9$                           & $75.2$                       \\
                                             & \DBE            & $\mathbf{0.03}$                  & $\mathbf{190.2}$                  & $22.0$                       \\
    \midrule \multirow{3}{*}{5$D$ AS}         & \SO             & $32.53$                          & $127.8$                           & $\mathbf{12.0}$              \\
                                             & \CBE            & $\mathbf{32.38}$                 & $\mathbf{72.2}$                   & $37.0$                       \\
                                             & \DBE            & $35.90$                          & $72.5$                            & $12.5$                       \\
    \midrule \multirow{3}{*}{10$D$ AS}        & \SO             & $\mathbf{59.69}$                 & $195.8$                           & $17.4$                       \\
                                             & \CBE            & $61.97$                          & $134.9$                           & $57.0$                       \\
                                             & \DBE            & $60.55$                          & $\mathbf{121.9}$                  & $\mathbf{17.0}$              \\
    \midrule \multirow{3}{*}{20$D$ AS}        & \SO             & $55.75$                          & $203.8$                           & $\mathbf{6.2}$               \\
                                             & \CBE            & $\mathbf{52.47}$                 & $183.2$                           & $104.0$                      \\
                                             & \DBE            & $61.16$                          & $\mathbf{151.6}$                  & $9.5$                        \\
    \midrule \multirow{3}{*}{40$D$ AS}        & \SO             & $319.00$                         & $130.5$                           & $\mathbf{3.0}$               \\
                                             & \CBE            & $388.20$                         & $139.5$                           & $102.2$                      \\
                                             & \DBE            & $\mathbf{309.60}$                & $\mathbf{108.1}$                  & $3.5$                        \\
    \midrule \multirow{3}{*}{5$D$ SE}         & \SO             & $0.09$                           & $115.5$                           & $\mathbf{12.5}$              \\
                                             & \CBE            & $\mathbf{0.02}$                  & $70.6$                            & $37.0$                       \\
                                             & \DBE            & $0.10$                           & $\mathbf{69.4}$                   & $\mathbf{12.5}$              \\
    \midrule \multirow{3}{*}{10$D$ SE}        & \SO             & $2.41$                           & $126.5$                           & $18.5$                       \\
                                             & \CBE            & $\mathbf{1.40}$                  & $89.2$                            & $75.0$                       \\
                                             & \DBE            & $2.37$                           & $\mathbf{77.4}$                   & $\mathbf{18.2}$              \\
    \midrule \multirow{3}{*}{20$D$ SE}        & \SO             & $17.67$                          & $118.1$                           & $\mathbf{14.1}$              \\
                                             & \CBE            & $21.78$                          & $95.2$                            & $76.0$                       \\
                                             & \DBE            & $\mathbf{17.56}$                 & $\mathbf{81.1}$                   & $14.8$                       \\
    \midrule \multirow{3}{*}{40$D$ SE}        & \SO             & $\mathbf{34.32}$                 & $293.9$                           & $\mathbf{23.0}$              \\
                                             & \CBE            & $52.46$                          & $231.9$                           & $122.5$                      \\
                                             & \DBE            & $45.02$                          & $\mathbf{184.6}$                  & $\mathbf{23.0}$              \\
    \midrule \multirow{3}{*}{5$D$ Rastrigin}  & \SO             & $10.94$                          & $92.7$                            & $\mathbf{11.0}$              \\
                                             & \CBE            & $11.28$                          & $62.0$                            & $35.0$                       \\
                                             & \DBE            & $\mathbf{10.85}$                 & $\mathbf{59.2}$                   & $\mathbf{11.0}$              \\
    \midrule \multirow{3}{*}{10$D$ Rastrigin} & \SO             & $28.24$                          & $96.5$                            & $\mathbf{14.8}$              \\
                                             & \CBE            & $24.71$                          & $73.7$                            & $54.2$                       \\
                                             & \DBE            & $\mathbf{23.89}$                 & $\mathbf{67.3}$                   & $15.0$                       \\
    \midrule \multirow{3}{*}{20$D$ Rastrigin} & \SO             & $59.08$                          & $129.4$                           & $21.5$                       \\
                                             & \CBE            & $57.04$                          & $95.4$                            & $68.8$                       \\
                                             & \DBE            & $\mathbf{55.81}$                 & $\mathbf{86.4}$                   & $\mathbf{20.0}$              \\
    \midrule \multirow{3}{*}{40$D$ Rastrigin} & \SO             & $88.31$                          & $169.2$                           & $26.5$                       \\
                                             & \CBE            & $100.36$                         & $120.6$                           & $87.0$                       \\
                                             & \DBE            & $\mathbf{76.48}$                 & $\mathbf{108.2}$                  & $\mathbf{25.8}$              \\
    \midrule
  \end{tabular}
  \caption{
    The benchmarking results of BO with $300$ trials using L-BFGS-B with the memory size of $m=10$ and $B=10$ restarts on the $D$-dimensional functions ($D=5,10,20,40$).
    We picked four objectives: Sphere, Attractive Sector (AS), Step Ellipsoidal (SE), and Rastrigin.
    The termination criterion of L-BFGS-B is $200$ iterations or $\|\nabla \alpha(\mathbf{x})\|_{\infty}\le 10^{-2}$.
    Each column shows the median over $20$ independent runs, each with a different random seed.
    \textbf{Best Value}: the minimum objective value among the 300 trials minus the best objective value over all runs.
    \textbf{Runtime (s)}: the total wall-clock time for BO.
    \textbf{Iters.}: the median of the L-BFGS-B iteration count over $300$ Trials $\times 10$ restarts. Lower is better.
  }
  \label{tab:bo_benchmark_all}
\end{table*}

Table~\ref{tab:bo_benchmark_all} provides the complete results for \emph{Sphere}, \emph{Attractive Sector}, \emph{Step Ellipsoidal}, and \emph{Rastrigin}.
Note that the setup follows Section~\ref{sec:experiments}.
The results show that \CBE\ degrades notably in higher dimensions both in solution quality and in optimizer iterations.
For example, \CBE\ yields larger \textbf{Best Value} in 40$D$ than \DBE\ by $+25.4\%$ on \emph{Attractive Sector}, $+16.5\%$ on \emph{Step Ellipsoidal}, $+31.3\%$ on \emph{Rastrigin}, and $+2.8\%$ on Sphere~\footnote{
  Percentages computed as $(\CBE-\DBE)/\DBE$ using Table~\ref{tab:bo_benchmark_all}.
}.
The iteration counts of \CBE\ also inflate sharply at $D{=}40$, requiring about $3.3{\times}\!\sim\!29{\times}$ more L-BFGS-B iterations than \DBE.
The largest gap is observed on \emph{Attractive Sector} (102.2 vs.\ 3.5).

By contrast, \DBE\ matches \SO\ in solution quality and iteration counts while achieving the best wall-clock time overall.
\DBE\ attains the shortest median runtime in almost all objective-dimension pairs (14/16), and is up to 1.76${\times}$ faster than \SO\ (5$D$ \emph{Attractive Sector}; $127.8$ seconds vs.\ $72.5$ seconds), and 1.29${\times}$ faster than \CBE (40$D$ \emph{Attractive Sector}; $139.5$ seconds vs.\ $108.1$ seconds).

    \fi
\end{document}